\documentclass[3p]{elsarticle}

\usepackage[utf8]{inputenc}
\usepackage[T1]{fontenc}
\usepackage{amssymb}
\usepackage{xcolor}
\usepackage{listings}
\usepackage{float}

\newfloat{lstfloat}{htbp}{lop}
\floatname{lstfloat}{Listing}

\usepackage{hyperref}
\hypersetup{
    colorlinks=true,
    linkcolor=blue,
    filecolor=magenta,      
    urlcolor=blue,
}
 
\usepackage{lmodern}

\usepackage{xcolor}

\definecolor{codegreen}{rgb}{0,0.6,0}
\definecolor{codegray}{rgb}{0.5,0.5,0.5}
\definecolor{codepurple}{rgb}{0.58,0,0.82}
\definecolor{backcolour}{rgb}{0.95,0.95,0.92}

\lstdefinestyle{mystyle}{
    commentstyle=\color{codegreen},
    keywordstyle=\color{magenta},
    numberstyle=\tiny\color{codegray},
    stringstyle=\color{codepurple},
    basicstyle=\ttfamily\footnotesize,
    breakatwhitespace=false,         
    breaklines=true,                 
    captionpos=b,                    
    keepspaces=true,                 
    numbers=left,                    
    numbersep=5pt,                  
    showspaces=false,                
    showstringspaces=false,
    showtabs=false,                  
    tabsize=2
}

\begin{document}
\begin{frontmatter}
    
\title{\textit{mlscorecheck}: Testing the consistency of reported performance scores and experiments in machine learning}

\author[1]{György Kovács}
\ead{gyuriofkovacs@gmail.com}

\author[2]{Attila Fazekas} 
\ead{attila.fazekas@inf.unideb.hu}

\affiliation[1]{organization={Analytical Minds Ltd.},
    addressline={Árpád str. 5}, 
    city={Beregsurany},
    postcode={4933}, 
    country={Hungary}}
    
\affiliation[2]{organization={Faculty of Informatics, University of Debrecen},
    addressline={Kassai str. 26}, 
    city={Debrecen},
    postcode={4028}, 
    country={Hungary}}

\begin{abstract}
    
    Addressing the reproducibility crisis in artificial intelligence through the validation of reported experimental results is a challenging task. It necessitates either the reimplementation of techniques or a meticulous assessment of papers for deviations from the scientific method and best statistical practices. To facilitate the validation of reported results, we have developed numerical techniques capable of identifying inconsistencies between reported performance scores and various experimental setups in machine learning problems, including binary/multiclass classification and regression. These consistency tests are integrated into the open-source package \emph{mlscorecheck}, which also provides specific test bundles designed to detect systematically recurring flaws in various fields, such as retina image processing and synthetic minority oversampling.    
\end{abstract}
    
    
\end{frontmatter}

\thispagestyle{empty}

\newpage

\noindent
\begin{tabular}{|l|p{6.5cm}|p{8cm}|}
\hline
\textbf{Nr.} & \textbf{Code metadata description} & 
\\
\hline
C1 & Current code version & 1.0.1 \\
\hline
C2 & Permanent link to code/repository used for this code version & \url{https://github.com/FalseNegativeLab/mlscorecheck} \\
\hline
C3  & Permanent link to Reproducible Capsule & \\
\hline
C4 & Legal Code License   & MIT \\
\hline
C5 & Code versioning system used & git \\
\hline
C6 & Software code languages, tools, and services used & Python; the \emph{pulp} package for linear programming\\
\hline
C7 & Compilation requirements, operating environments \& dependencies & Operating systems: Windows, Linux, MacOS; Python 3.9+; the Python packages \emph{numpy}, \emph{scipy}, \emph{scikit-learn}, \emph{pulp}\\
\hline
C8 & If available Link to developer documentation/manual & \url{https://mlscorecheck.readthedocs.io/en/latest/}\\
\hline
C9 & Support email for questions & \href{mailto:gyuriofkovacs@gmail.com}{gyuriofkovacs@gmail.com}\\
\hline
\end{tabular}\\
\vskip0.5cm

\newpage

\section{Introduction}

Whether researchers are conducting basic research or refining practical applications, they routinely evaluate and rank the performance of machine learning techniques using quantitative metrics and scores \cite{scores}. However, several factors, including improperly conducted experiments \cite{ehg}, the misuse of statistics \cite{staterrors}, data leakage \cite{leakage}, and typographical errors, can compromise the replicability of reported performance scores in independent experiments. This phenomenon contributes to what is often referred to as the \emph{reproducibility crisis} in machine learning research \cite{reprcrisis}. Moreover, the problem of unreliable performance scores is intensified by the \emph{publication bias} \cite{publicationbias}, which has the potential to derail entire research fields \cite{ehg}.

Identifying unreliable scores is a time-consuming process, often requiring the reimplementation of published methods \cite{ehg}. It would be desirable to have numerical techniques that facilitate the quick validation of reported results. These techniques could utilize that performance scores calculated in a particular experiment cannot take arbitrary values independently. The domain and interrelation of the scores are limited by the experimental setup and the mathematical relations between them.

Utilizing the interrelation of performance scores in binary classification, a technique called \emph{DConfusion} was proposed in \cite{dconfusion, errorsml} to reconstruct certain characteristics 
(but not the exact entries) 
of the confusion matrix from reported scores. 
Independently, in \cite{vesselsegm, vessel}, we proposed similar techniques for the consistency testing of reported performance scores in the field of retinal vessel segmentation, successfully identifying systematic flaws affecting more than 100 papers in the field \cite{vessel}. Our approach differs from \emph{DConfusion} in several ways: (a) unlike \emph{DConfusion}, which supports only a limited set of scores, the proposed technique supports the majority of performance scores used in the literature; (b) \emph{DConfusion} neglects the impact of aggregations, whereas the proposed method addresses the aggregation of scores with mathematical rigor; (c) due to the omission of aggregations and rounding, \emph{DConfusion} might produce false alarms of inconsistency, whereas the inconsistencies identified by the proposed technique are certain; (d) DConfusion supports only binary classification, whereas the proposed technique covers all major tasks of supervised learning (binary/multiclass classification, regression).

In this paper, we introduce (Section \ref{sec:brief}) and illustrate (Section \ref{sec:main}) the Python package \emph{mlscorecheck}, generalizing the techniques proposed in \cite{vessel} to test the consistency of reported performance scores in various experimental setups. Potential impacts are discussed in Section \ref{sec:impact} and conclusions are drawn in Section \ref{sec:summary}.

\section{The numerical methods of consistency testing}
\label{sec:brief}

The consistency testing of reported performance scores requires three inputs: (a) a set of reported score values; (b) the numerical uncertainty of the reported figures (when scores are truncated to $k$ decimal places, the real values can be assumed within the radius of $10^{-k}$); and (c) the description of the experimental setup (the statistics of the dataset(s), the cross-validation scheme, etc.). The consistency tests rigorously answer the question: \emph{could the reported scores be yielded from the described experiment?} There are three main types of scenarios that necessitate dedicated testing methodologies:
\begin{enumerate}
\item When \emph{performance scores are calculated from confusion matrices \cite{scores} directly}, exhaustive enumeration is employed and expedited by interval computing. This test is applicable in binary and multiclass classification when there is only one testset or \emph{score of means}/\emph{micro-average} \cite{multiclass} aggregation is used. These tests support 20 commonly used performance scores, including accuracy, F$_1$ and more \cite{scores}.
\item When the reported \emph{performance scores are the averages of fold/dataset-level scores} in classification, exhaustive enumeration becomes intractable. In such cases (typically k-fold cross-validation scenarios with \emph{mean of scores}/\emph{macro-average} \cite{multiclass} aggregation), linear integer programming techniques are used for testing.
\item In the absence of confusion matrices \emph{in regression problems}, the test relies on the necessary mathematical relation of scores or terms of the scores. The consistency tests for regression currently support the \emph{mean average error}, \emph{mean squared error} and the $r^2$ scores.
\end{enumerate}

Analogously to statistical testing, in each test, the null-hypothesis is that the configuration is consistent, and whenever inconsistency is identified, it provides evidence against the null-hypothesis, but being numerical rather than statistical, the evidence is indisputable.
For a detailed mathematical description of the tests and the list of supported scenarios, refer to \cite{vessel, fazekas2023testing} and documentation \url{https://mlscorecheck.readthedocs.io/en/latest/}.

\section{Main functionalities}
\label{sec:main}

The package provides two main functionalities: (a) high-level functions for testing the consistency of performance scores and experimental setups in general, and (b) test bundles dedicated to certain widely researched problems. The following sections illustrate these two functionalities.

\subsection{General testing functionalities}

The package offers tests for various experimental setups, including the use of k-fold cross-validation with known/unknown fold configurations, the use of various aggregations (micro-/macro-averages), and potentially multiple datasets. For the correct outcome, it is crucial to specify all presumed details of the experiments accurately. This leads us to the design principle that dedicated functions have been implemented for the various experiments, thereby guiding the user to carefully consider all important details when choosing the test to be executed. In particular, altogether, there are 17 test functions for binary classification (10), multiclass classification (6), and regression (1) scenarios.

For illustration, in a general scenario, we consider an experiment where the performance scores accuracy, sensitivity and F$_1$ are reported on a testset consisting of $p=100$ positive and $n=1000$ negative samples. If the experiment leads to 81 true positive and 850 true negative samples, and the performance scores are rounded to 4 decimal places, the reported scores become $acc=0.8464$, $sens=0.81$ and $F_1 = 0.4894$. In general, the real number of true positive and true negative samples is unknown, and we want to know if the reported scores could be the outcome of the presumed experiment. Using the \emph{mlscorecheck} package to carry out this consistency testing is as easy as shown in Listing \ref{list0}: the specification of the experiment happens through the choice of the test function; the description of the testset and the reported scores are given; and the numerical uncertainty of $0.0001$ is inferred from the performance scores assuming truncation to 4 decimal places. Among many details of the test, the 'inconsistency' flag of the outcome indicates that inconsistency was not identified (which is natural in this example as the scores are calculated from a real choice of true positives and true negatives). 

\begin{lstfloat}[t]
\begin{lstlisting}[language=Python, caption={Consistency testing in the case of testset, with no k-fold cross-validation}, basicstyle=\scriptsize, label=list0
    %,numbers=none
]
from mlscorecheck.check.binary import check_1_testset_no_kfold

result = check_1_testset_no_kfold(testset={'p': 100, 'n': 1000},
                                  scores={'acc': 0.8464, 'sens': 0.81, 'f1': 0.4894},
                                  eps=1e-4)
result['inconsistency']
#False
\end{lstlisting}
\end{lstfloat}

Contrarily, if one of the scores is slightly adjusted (for example, accuracy changed to 0.8474) or the number of positive samples is incorrectly assumed to be 110 in the testset, the test indicates that inconsistency has been identified: there is no such true positive and true negative combination that could lead to the reported scores in the presumed experiment.

All high-level test functions follow the same philosophy. One selects the test function suitable for the experiment (the number of datasets involved, use of k-fold schemes, mode of aggregation), provides the available data (the statistics of the dataset(s) and the reported scores), and the test returns if inconsistency has been identified.

\subsection{The test bundles}

To facilitate the use of the package to test the scores reported in widely researched problems, we provide predefined experiment specifications for certain fields and datasets.
The test bundles cover three fields of machine learning research in medicine, with experiment specifications for a total of 10 publicly available datasets (7 for retinal image processing \cite{vessel}, 2 for skin lesion classification \cite{skinsurvey}, and 1 for preterm delivery prediction from electrohysterogram signals \cite{tpehg}).

The use of the test bundles is illustrated by an example in the field of skin lesion classification. Suppose there is a study reporting the performance scores accuracy (0.7916), sensitivity (0.2933) and specificity (0.9145) for the skin lesion classification dataset ISIC2016 \cite{isic2016}. Since the experimental setup is predefined in the package, consistency testing can be carried out without specifying the experiment and the dataset, as illustrated in Listing \ref{list2}. The results show that inconsistency was not found, however, a slight adjustment (accuracy changed oto 0.7926) would lead to inconsistencies.

\begin{lstfloat}[t]
\begin{lstlisting}[language=Python, caption={Consistency testing of scores reported for the ISIC2016 dataset}, basicstyle=\scriptsize, label=list2
    %, numbers=none
]
from mlscorecheck.check.bundles.skinlesion import check_isic2016

result = check_isic2016(scores={'acc': 0.7916, 'sens': 0.2933, 'spec': 0.9145}, 
                         eps=1e-4)
result['inconsistency']
#False
\end{lstlisting}
\end{lstfloat}


\section{Impact overview}
\label{sec:impact}

In this section, we briefly discuss three fields where inconsistencies have been identified using the methods in the package and explore potential applications.

Earlier, we applied the developed methods to analyze over 100 papers in the field of retinal vessel segmentation \cite{vessel} and highlighted a systematic flaw in the evaluation methodologies that biases the ranking of algorithms in almost all papers. This flaw, related to the circular field of view of retinal images, potentially affects all areas of retinal image processing, such as exudate segmentation \cite{exu} and optic disk segmentation \cite{od}. In a manner similar to \cite{vessel}, the tools in the package are suitable for the meta-analysis of all these problems. The test bundles prepared in the package support more than 15 experiments on 7 public datasets of retinal imaging.

The research \cite{ehg} identified 11 papers with overly optimistic results related to the prediction of preterm delivery from electrohysterogram signals. The flawed predictions resulted from the improper use of synthetic minority oversampling \cite{smote}, which can be detected solely from the reported scores \cite{fazekas2023testing} using the \emph{mlscorecheck} package, thereby avoiding the laborious reimplementation of 11 techniques as done in \cite{ehg}. This example also shows that the techniques are suitable to validate numerical results published in the field of widely researched field of imbalanced learning \cite{imblearn}.

As a final example, the methods implemented in the package were used for a brief meta-analysis of the field of skin lesion classification and facilitated the identification of highly cited studies with performance scores that could not be the outcome of the claimed experimental setups \cite{fazekas2023testing}.

\section{Summary}
\label{sec:summary}

In this paper, we have provided a brief introduction to the main functionalities of the Python package \emph{mlscorecheck}, which is dedicated to the consistency testing of machine learning performance scores (sections \ref{sec:brief} and \ref{sec:main}). At the time of writing, the package supports a total of 17 general experimental setups for binary/multiclass classification and regression scenarios. In addition to the general functionalities, the package is equipped with more than 20 test bundles for 10 widely used, open-access machine learning and image processing datasets in medicine.


Regarding limitations, we believe that the package supports many of the commonly used experimental setups. However, support for further types of cross-validation schemes (for example, the repeated hold-out validation which is commonly used in medicine) is to be added in future releases. 

Based on the successful applications \cite{vessel,fazekas2023testing} and considering the numerous potential future use cases (Section \ref{sec:impact}), we believe that the package can become an effective tool for the meta-analysis of already published and future machine learning research, contributing to the enhancement of reproducibility in the field.


The \emph{mlscorecheck} package is available on GitHub at \url{https://github.com/FalseNegativeLab/mlscorecheck}, with detailed documentation accessible at \url{https://mlscorecheck.readthedocs.io/en/latest/}, discussing and illustrating various applications and use cases. The quality of the implementation is ensured by more than 22k unit, integration, and functional tests, adhering to the best practices of open-source scientific software development.

Finally, as a call for contributions, experts from various fields are welcome to submit test specifications to facilitate the numerical validation of performance score reported for publicly available datasets in various fields.



\bibliographystyle{abbrv}
\bibliography{references}



\end{document}